\documentclass[10pt,twocolumn,letterpaper]{article}

\usepackage{iccv}
\usepackage{times}
\usepackage{epsfig}
\usepackage{graphicx}
\usepackage{amsmath}
\usepackage{amssymb}
\usepackage{pifont}
\newcommand{\cmark}{\ding{51}}%
\newcommand{\xmark}{\ding{55}}%

\usepackage{algorithm}
\usepackage[noend]{algpseudocode}
\usepackage[pagebackref=true,breaklinks=true,letterpaper=true,colorlinks,bookmarks=false]{hyperref}

\iccvfinalcopy % *** Uncomment this line for the final submission

\def\iccvPaperID{5136} % *** Enter the ICCV Paper ID here

\ificcvfinal\fi

\begin{document}

%%%%%%%%% TITLE
\title{CapsuleVOS: Semi-Supervised Video Object Segmentation Using Capsule Routing}

\author{Kevin Duarte\\
%Center for Research in Computer Vision\\
%University of Central Florida\\
%Orlando, FL 32816\\
{\tt\small kevin\_duarte@knights.ucf.edu}
\and
Yogesh S Rawat\\
%Center for Research in Computer Vision\\
%University of Central Florida\\
%Orlando, FL 32816\\
{\tt\small yogesh@crcv.ucf.edu}
\and
Mubarak Shah\\
% Center for Research in Computer Vision\\
% University of Central Florida\\
% Orlando, FL 32816\\
{\tt\small shah@crcv.ucf.edu}
\and
% Mubarak Shah\\
Center for Research in Computer Vision\\
University of Central Florida\\
Orlando, FL 32816\\
}

\maketitle
%\thispagestyle{empty}

%%%%%%%%% ABSTRACT
\begin{abstract}
%In this work we propose a capsule network for semi-supervised video object segmentation. 
%limitations of existing methods, requirement of flow and frame based..
%novel routing mechanism for attention based capsule selection...
%We propose a novel zooming module which helps the network focus on small object of interest...
%The framework also utilizes a novel memory module based on recurrent network which helps in tracking objects when they move out of frame in the video...
%the network can be trained end-to-end...
%demonstrated the effectiveness of the network on two different video object segmentaiton datasets...
%state of the art results...

In this work we propose a capsule-based approach for semi-supervised video object segmentation. Current video object segmentation methods are frame-based and often require optical flow to capture temporal consistency across frames which can be difficult to compute. To this end, we propose a video based capsule network, CapsuleVOS, which can segment several frames at once conditioned on a reference frame and segmentation mask. This conditioning is performed through a novel routing algorithm for attention-based efficient capsule selection. We address two challenging issues in video object segmentation: 1) segmentation of small objects and 2) occlusion of objects across time. The issue of segmenting small objects is addressed with a zooming module which allows the network to process small spatial regions of the video. Apart from this, the framework utilizes a novel memory module based on recurrent networks which helps in tracking objects when they move out of frame or are occluded. The network is trained end-to-end and we demonstrate its effectiveness on two benchmark video object segmentation datasets; it outperforms current offline approaches on the Youtube-VOS dataset while having a run-time that is almost twice as fast as competing methods. The code is publicly available at \href{https://github.com/KevinDuarte/CapsuleVOS}{https://github.com/KevinDuarte/CapsuleVOS}.
%\MS {The abstract is pretty short and low key, not much wow factor! It will be good strengthen it to excite the reviewers to read it further with interest.}
%\MS {Abstract needs to be improve signficantly; which can be done at the end}
\end{abstract}

%%%%%%%%% BODY TEXT
\section{Introduction}

Semi-supervised video object segmentation aims to segment objects in a video,  given their segmentation masks for the first frame. This is a challenging problem because of issues like occlusion, changes in object appearance over time, motion blur, fast motions, and scale variations of different objects. Deep learning approaches have achieved impressive results and the recent release of the Youtube-VOS dataset \cite{xu2018youtube2} has allowed for the training and evaluation of new methods on a wider variety of videos and objects.

The majority of current approaches can be divided into two categories. The first are detection-based methods \cite{ caelles2017one, chen2018blazingly, hu2018videomatch} that learn representations of the object segmented in the first frame and attempt to perform the pixel-wise detection of this object in future frames; the second is propagation-based methods \cite{ci2018video, hu2018motion, tokmakov2017learning, xiao2018monet, xu2018youtube} that formulate the task as a tracking problem and attempt to propagate the mask to fit the object over time. The first set of methods tends to segment single frames independently and rarely employ temporal information, while the later set segments single frames sequentially and makes use of temporal information, usually in the form of optical flow or RNNs. There has been some work on hybrid methods, that attempt to unify both approaches \cite{wug2018fast, li2018video, yang2018efficient}.

We propose a hybrid method that makes use of a {\em video capsule network} to segment a video conditioned on the segmented object in the first frame. A capsule is a group of neurons that represents an object, or part of an object. Layers in capsule networks undergo a routing-by-agreement algorithm that finds similarities between these capsules, and allow for the modeling of part-to-whole relationships. Capsule networks have performed well in image classification \cite{sabour2017dynamic, hinton2018matrix}, and have shown outstanding results in various segmentation tasks \cite{lalonde2018capsules, duarte2018videocapsulenet}. In this paper, we leverage the segmentation ability of capsule networks and the ability of the routing algorithm to find similarity between capsules for the task of semi-supervised video object segmentation.

Our video capsule network, CapsuleVOS, contains two branches: a video branch and a frame branch. The video branch processes several frames at once and produces a set of video capsules. This allows the network learn temporal/motion information without the reliance of optical flow. The frame branch processes the first frame and object segmentation and generates a set of frame capsules, which model the object of interest. The frame branch makes use of a recurrent memory module that allows the network to overcome issues like occlusion or objects exiting the scene.

Both sets of capsules are then passed through our novel {\em attention-routing procedure} which allows the frame capsules to condition the video capsules. Through this routing algorithm, our network learns where the object of interest is within the video clip, allowing the network to {\em segment multiple frames simultaneously}.

Moreover, our method makes use of a parametrized zooming module which allows the network to focus on regions of the frame which are relevant to the object of interest. This module allows for the segmentation of smaller objects, which can easily be lost when resizing frames to lower spatial dimensions.

We make the following contributions in this work,
\begin{itemize}
    \item We present a novel capsule network for the task of video object segmentation that achieves state-of-the-art results on the largest video segmentation dataset.
    \item We propose a novel attention based EM routing algorithm to condition capsules based on an input segmentation.
    \item The proposed network contains integrated zooming module and memory module, which we show through experimental results to be effective for segmenting small and occluded objects in the video.
\end{itemize}

\section{Related Work}

\paragraph{Semi-supervised video object segmentation:} 
Earlier works in video object segmentation used hand-crafted features based on appearance, boundary and optical flow \cite{brox2010object, faktor2014video, jain2014supervoxel, Nagaraja_2015_ICCV, papazoglou2013fast}. The availability of large-scale video object segmentation datasets \cite{pont20172017, xu2018youtube2} enabled us to explore deep learning methods for this problem. Most of the early works are mainly motivated by the image segmentation methods \cite{chen2018deeplab, xu2016deep, long2015fully}. These works \cite{caelles2017one, cheng2017segflow, jain2017fusionseg, perazzi2016benchmark, yang2018efficient} lack the integration of sequential modelling which is important from video perspective. In some of these works, the temporal consistency is achieved by taking a guidance from the predicted mask of the previous frame \cite{hu2017maskrnn, perazzi2016benchmark, yang2018efficient}. The majority of recent works also utilize online learning \cite{caelles2017one} in which the segmentation networks are fine-tune on the first frame of each test video - this greatly improves segmentation results at the expense of inference speed. %\MS {Do you want to put a sentence here to explain what does online learning do, I am not sure if most reviewers will know about this.}
%changed - Kevin

Several recent works have utilized recurrent units to learn the evolution of objects over time. The authors in \cite{tokmakov2017learning}  use a ConvGRU to combine the outputs of pretrained appearance and a motion networks and generate a final segmentation. Similarly, the authors in \cite{xu2018youtube} propose a ConvLSTM sequence-to-sequence model that learns to generate segmentations from sequences of frames. Ventura \etal \cite{ventura2019rvos} also use a ConvLSTM for recurrence in both the temporal domain (between frames) and the spatial domain (between object instances within each frame).  Our use of a recurrent memory unit  differs from these methods in that we do not generate segmentations directly from the features generated by the ConvLSTM,  but rather condition a segmentation network based on these features. 

Segmentation of small objects is challenging and zooming in on regions of the frame has been explored to overcome this problem. The authors in  \cite{ci2018video} demonstrated the effectiveness of processing only a tight region around the foreground object. Although this allows for improved segmentations, it assumes the object moves smoothly within the video - in cases of large motions, this may fail. Our approach can handle this issue, since our network learns the extent to which it must zoom in on the object of interest, allowing the network to learn these cases where large motions occur. The work in \cite{cheng2018fast} performs segmentation by tracking parts - their network zooms in on and processes each part of the object separately. This requires multiple passes through their segmentation model, instead of having a single segmentation of the whole object.%\MS{Why it is bad?}

\paragraph{Capsule networks:} 
The idea of capsules was first introduced in \cite{hinton2011transforming}, and they were popularized in \cite{sabour2017dynamic}, where dynamic routing for capsules was proposed. This was further extended in \cite{hinton2018matrix}, where a more effective EM routing algorithm was introduced. Recently, capsule networks have shown state-of-the-art results for human action localization in video \cite{duarte2018videocapsulenet}, object segmentation in medical images \cite{lalonde2018capsules}, and text classification \cite{zhao2018investigating}. In this work, we propose a capsule based network for video object segmentation where we introduce a novel attention based EM routing which can be used as a conditioning mechanism for capsules. %\MS{It will be better to put motivation of this new routing, say why the old routing has problems?}

\section{Our Approach}
\begin{figure}[t]
\centering
%\fbox{\rule{0pt}{2in} 
%\includegraphics[width=0.9\linewidth]{Figures/Duag1_new2.PNG}
\includegraphics[width=0.9\linewidth]{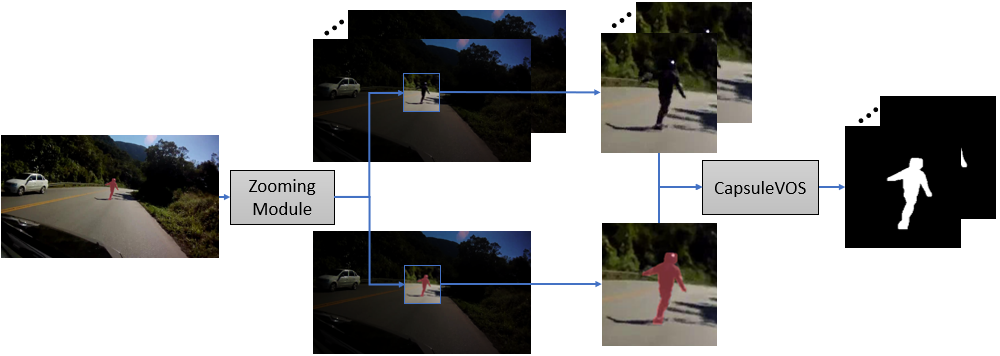}
   \caption{Zooming Module. Given the high-resolution first frame and segmentation mask, the zooming module outputs a bounding box around the object of interest. This bounding box is used to zoom in on the object in the video clip along with the first frame and segmentation mask, which are resized and passed into the CapsuleVOS network.}
\label{fig:zooming}
\end{figure}

We propose an end-to-end trained network that segments an object throughout an entire video clip  when given the object's segmentation mask for the first frame. This network contains two modules, depicted in Figures \ref{fig:capsvos} and \ref{fig:zooming}: a frame-conditioned video capsule network, CapsuleVOS, which segments a short video clip (8 frames) based on the object segmentation in the first frame, and a zooming module, which refines the spatial area processed by the capsule network. Section \ref{section:conditioning} explains how we leverage capsules for the task of video object segmentation, with our attention-based routing algorithm. We then describe the CapsuleVOS architecture and the zooming module in sections \ref{section:capsvos} and \ref{section:zooming} respectively. This is followed by the objective function used to train this network in section \ref{section:objective}.

\begin{figure*}[t]
\centering
%\fbox{\rule{0pt}{2in} 
\includegraphics[width=0.9\linewidth]{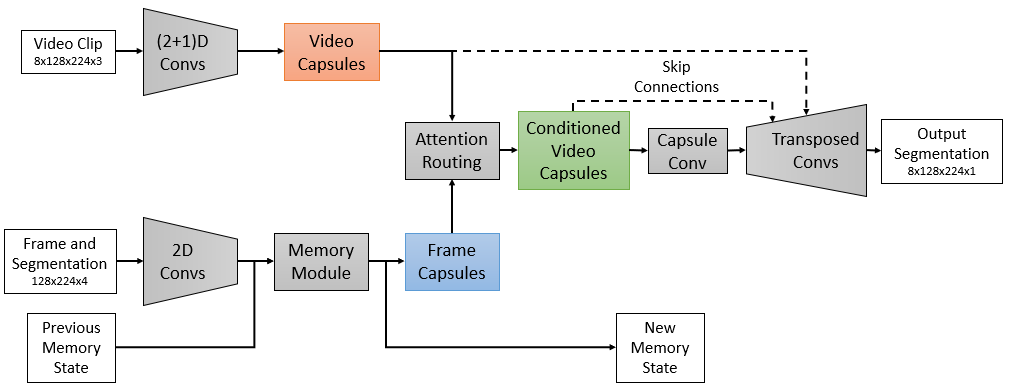}
   \caption{CapsuleVOS Architecture. The network is given the low resolution video clip and the segmented object in the first frame, and generates the foreground segmentations for all frames of the clip. The memory module consists of a ConvLSTM and allows the network to overcome issues like occlusion and objects leaving the frame. The previous and new memory states are the hidden and cell states of the ConvLSTM for time steps $t$ and $t-1$ respectively. The new memory state is passed to the memory module for the following video clip.}
\label{fig:capsvos}
\end{figure*} 

\subsection{Conditioning with Capsules} \label{section:conditioning}

Capsules are groups of neurons that represent different entities or objects. In this work, we employ the version of capsules described in \cite{hinton2018matrix}, which have a logistic unit (an activation denoted by $a$) representing the presence of the entity and a $4 \times 4$ pose matrix (denoted by $M$) which contains the properties of the entity. Capsules in one layer vote for the pose matrices of many capsules in the following layer and an iterative EM routing algorithm finds the agreement between the votes to create the set of capsules in the next layer. For a more comprehensive understanding of capsules, and the intuition behind them, we suggest reading \cite{sabour2017dynamic, hinton2018matrix}.

We view capsule networks' ability to model entities and find agreement between entities as an ideal mechanism to accomplish the semi-supervised video object segmentation task. A given video may contain several objects and the reference segmentation mask specifies the object which must be segmented. If we extract a set of capsules from both the video and the reference frame with a segmentation mask, then the former set (video capsules) models all objects within the video, while the latter set (frame capsules) represents the object of interest. Then, to obtain the object of interest throughout the video, one only needs to filter out all video capsules that are dissimilar to the frame capsules; in other words, an agreement, or similarity, between the video capsules and frame capsules would result in the set of video capsules that represent the object that must be segmented. Although the original EM routing algorithm works well for finding agreement within a set of capsules, it can not explicitly find agreement between two sets of capsules. For this reason, we propose an attention-based routing algorithm which finds the agreement between two sets of capsules.

%When describing our capsule conditioning method, 
Here, we use the query, key, value terminology found in \cite{vaswani2017attention}, as our conditioning algorithm takes inspiration from this attention mechanism. From a video clip we extract a set of the video capsules $M_i^\mathcal{V}, a_i^\mathcal{V}$, indexed by $i$; from a reference frame and segmentation mask, we extract a set of frame capsules $M_k^\mathcal{F}, a_k^\mathcal{F}$, indexed by $k$. The key-value pairs are votes from the video capsules for the following layers' capsules while the query is the set of votes from the frame capsules. These votes are calculated as follows:
\begin{equation}
    \begin{aligned}
    V_{ij}^{k} = M_i^\mathcal{V} W_{ij}^{k}\\
    V_{ij}^{v} = M_i^\mathcal{V} W_{ij}^{v}\\
    V_{kj}^{q} = M_k^\mathcal{F} W_{kj}^{q}
    \end{aligned}
\end{equation}
where $W_{ij}^{k}$, $W_{ij}^{v}$, and $W_{kj}^{q}$ are learned weight matrices. The superscripts $k$, $v$, and $q$ correspond to the key, value, and query respectively.

Once these votes are obtained, the EM routing operation is performed for the frame capsule (query) votes. This results in a set of higher-level capsules $M_j^q, a_j^q$, which represents the object, or parts of the object, in the reference segmentation mask. To find the similarity, or agreement, between the video capsules and the frame capsules, we  measure the Euclidean distance between the key votes ($V_{ij}^{k}$) and their corresponding higher-level query capsule:
\begin{equation}
    D_{ij} = \sum_h \left[\left( M_j^q - V_{ij}^k\right)^2 \right]^h,
\end{equation}
where $h$ denotes the dimensions of the vote and pose matrices.

This distance is used to compute an assignment coefficient %\MS {can you come with some other term for this}, 
\begin{equation}
    R_{ij}^v = \frac{e^{-D_{ij}}}{\sum_j e^{-D_{ij}}}. % could write out the softmax instead of having the text - Kevin
\end{equation}
The assignment coefficient, $R_{ij}^v$, determines the amount of information the $i^\text{th}$ video capsule sends to the $j^\text{th}$ higher-level capsule. If the distance, $D_{ij}$, is large, then the $i^\text{th}$ video capsule does not contain information pertaining the the object represented by the $j^\text{th}$ higher-level capsule, so its corresponding assignment coefficient is close to $0$, and it sends less information to that higher-level capsule; conversely, a small distance leads to a large assignment coefficient, resulting in more information being sent.

We obtain the conditioned set of video capsules by performing the M-step of the EM routing algorithm using the value votes ($V_{ij}^{v}$) and the video capsules' assignment coefficients. The result is a set of higher-level video capsules, $M_j^v, a_j^v$, that receive information from lower-level video capsules which agree with the frame capsules. This procedure of conditioning with capsules is described in Algorithm \ref{alg:routing}.
\begin{algorithm}
\caption{This routing algorithm returns the activations and pose matrices of the capsules in layer $L+1$ when given the activations and poses of layer $L$ (the video capsules and frame capsules). The indices $i$ and $j$ refer to the capsule types in layer $L$ and $L+1$ respectively. The index $h$ refers to the dimensions of the vote or pose matrices. The $\textsc{EM Routing}$ and $\textsc{M-Step}$ functions referenced are those defined in \cite{hinton2018matrix}.}\label{alg:routing}
\begin{algorithmic}[1]
\Procedure{AttRouting}{$M^\mathcal{V},\, a^\mathcal{V},\, M^\mathcal{F},\, a^\mathcal{F}$}
\State $V^{v} \gets M^\mathcal{V} W^{v}$
\State $V^{k} \gets M^\mathcal{V} W^{k}$
\State $V^{q} \gets M^\mathcal{F} W^{q}$
\State $a^q, M^q\gets\textsc{EM Routing}(a^\mathcal{F}, V^{q})$
\State $D_{ij} \gets \sum_h\left[\left( M_i^q - V_{ij}^k \right)^2\right]^h$ \Comment{For each $i$ and $j$}
\State $R_{ij}^v \gets \frac{e^{-D_{ij}}}{\sum_j e^{-D_{ij}}}$  \Comment{For each $i$}
\State $a_j^v, M_j^v \gets \textsc{M-Step}(a^\mathcal{V}, R^v, V^{v}, j)$ \Comment{For each $j$}
\State \textbf{return} $a^v, M^v$
\EndProcedure
\end{algorithmic}
\end{algorithm}

\iffalse
\begin{figure*}[t]
\centering
%\fbox{\rule{0pt}{2in} 
\includegraphics[width=0.9\linewidth]{Figures/Diag2.PNG}
   \caption{CapsuleVOS Architecture. The network is given the low resolution video clip and the segmented object in the first frame, and generates the foreground segmentations for all frames of the clip. The memory module consists of a ConvLSTM and allows the network to overcome issues like occlusion and objects leaving the frame. The previous and new memory states are the hidden and cell states of the ConvLSTM for time steps $t$ and $t-1$ respectively. The new memory state is passed to the memory module for the following video clip.}
\label{fig:capsvos}
\end{figure*} 
\fi

%\section{Our Approach}
%We propose an end-to-end trained network that segments an object throughout an entire video clip  when given the object's segmentation mask for the first frame. This network, depicted in Figure \ref{fig:network}, contains two modules: a frame-conditioned video capsule network, CapsuleVOS, which segments a short video clip (8 frames) based on the object segmentation in the first frame, and a zooming module, which refines the spatial area processed by the capsule network.

%\MS{Should conditioning with capsules section be under Our approach?}

\subsection{CapsuleVOS Architecture} \label{section:capsvos}
% could rename frame branch to conditioning branch - Kevin
The CapsuleVOS network segments 8 frames based on the segmentation mask of the first frame. It contains two branches - the video branch and the frame branch - and each creates sets of capsules. The video capsules are conditioned on the frame capsules, to produce a new set of conditioned capsules. These are followed by a convolutional capsule layer and a series of transposed convolutions to generate a segmentation map for all 8 frames.

The video branch passes the 8 RGB frames of size $128 \times 224$ through 6 (2+1)D convolutions \cite{tran2018closer} to obtain feature maps of size $8 \times 32 \times 56 \times 512$. The video capsules are composed of 12 capsule types, which are obtained by passing the feature maps to strided $3 \times 3 \times 3$ convolution operations.

The frame branch concatenates the first frame and the segmentation mask (each of size $128 \times 224$) and passes them through 4 2D convolutions. This is followed by the memory module, which consists of a ConvLSTM \cite{xingjian2015convolutional} layer that allows the frame branch to maintain information which might be lost in cases of occlusion or objects leaving the frame. The ConvLSTM produces a set of features of shape $32 \times 56 \times 128$ which are transformed into the frame capsules through a strided $3 \times 3$ convolution operation. The frame capsules, which are composed of 8 capsule types, are then tiled 8 times to match the temporal dimension of the video capsules.

Once the video and frame capsules have been formed, we perform capsule conditioning as described in Section \ref{section:conditioning}, which results in a set of 16 capsule types. This is followed by a convolutional capsule layer that has 16 capsule types. All routing operations make use of capsule pooling \cite{duarte2018videocapsulenet} to reduce network's memory consumption. 

To obtain a foreground segmentation mask from this capsule representations we flatten the capsules' pose matrices and pass them to a decoder composed of strided transposed convolutions. Skip connections from the video capsules and conditioned capsules are used to maintain spatiotemporal information which is lost from striding. The result of this decoder is 8 frames of binary segmentations corresponding to the object of interest.

\subsection{Zooming Module} \label{section:zooming}
% could move the first two sentences to the related works - Kevin
The zooming module is given the high-resolution first frame and the object of interest segmentation mask, and it outputs the bounding box containing the spatial region which our segmentation network will process. Since our segmentation network processes 8 frames at a time, the predicted bounding box must be large enough to contain the object of interest in all 8 frames, but not too large as to contain extraneous information not necessary for segmentation.

The input for the zooming module is a high-resolution frame ($512 \times 896$) and the high-resolution binary object segmentation mask. These are passed through a series of strided 2D convolutional layers, a LSTM layer, and a fully-connected layer which outputs two values, $\hat{b}_h$ and $\hat{b}_w$, representing the height and the width of the bounding box centered on the object of interest. The LSTM layer allows the network to learn from motion information from previous time steps, resulting in larger bounding boxes for objects with more motion, and tighter bounding boxes for objects with relatively little motion. Once the bounding box is obtained, the network extracts this region from the high-resolution segmentation mask and the next 8 frames of the high-resolution video; these are then resized to $128 \times 224$ and passed to CapsuleVOS. 

\begin{figure*}[ht!]
\begin{center}
    
\includegraphics[width=1.0\linewidth]{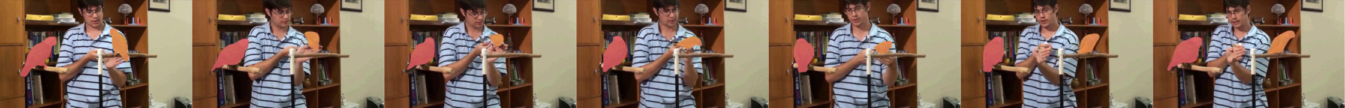}
\includegraphics[width=1.0\linewidth]{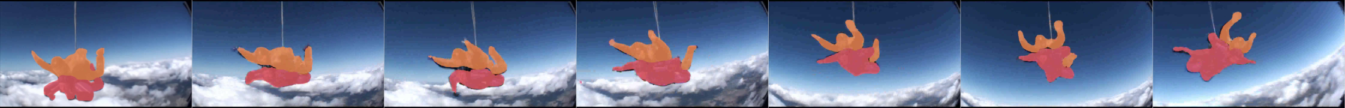}
\includegraphics[width=1.0\linewidth]{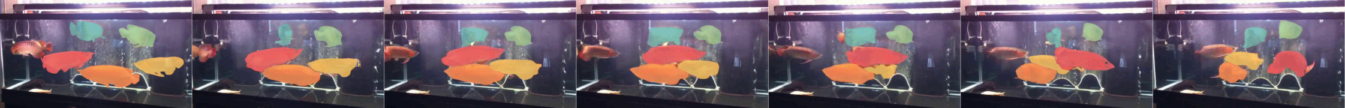}
\includegraphics[width=1.0\linewidth]{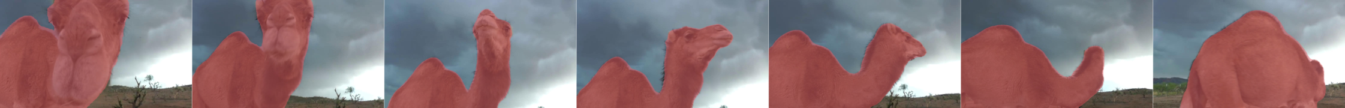}
\includegraphics[width=1.0\linewidth]{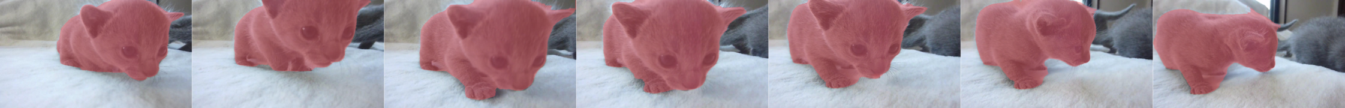}

\end{center}
   \caption{Qualitative results showing object segmentations on videos from the Youtube-VOS validation set. The first three rows contain examples in which multiple instances of objects are present within the video; the later two show how our network is able to finely segment larger objects.}
\label{fig:qualitative-good}
\end{figure*}

\subsection{Objective Function} \label{section:objective}
For each pixel $i$ in the video, we have ground-truth segmentations $y_i \in \{ 0, 1\}$ and our network predicts $\hat{y}_i \in [0, 1]$. We use both binary cross-entropy
\begin{equation}
    L_s = -\frac{1}{N}\sum_{i=1}^N y_i \log \left(\hat{y}_i \right) + (1-y_i) \log \left(1-\hat{y}_i \right),
\end{equation}
and the dice loss \cite{milletari2016v}
\begin{equation}
    L_D = 1-\frac{\sum_{i=1}^N {\hat{y}_i y_i + \epsilon}}{\sum_{i=1}^N {\hat{y}_i+ y_i + \epsilon}} - \frac{\sum_{i=1}^N {\left(1-\hat{y}_i \right)\left(1- y_i \right) + \epsilon}}{\sum_{i=1}^N {2- \hat{y}_i - y_i + \epsilon}},
\end{equation}
to train the network for segmentation. The $\epsilon$ term is a small value to ensure stability of the loss. We use this second segmentation loss because video object segmentation methods are evaluated using region similarity, or intersection-over-union (IoU), and the dice loss directly maximizes this metric.

We train the zooming module by computing the L2 loss between the ground-truth bounding box height and width ($b_h$ and $b_w$) and the predicted height and width ($\hat{b}_h$ and $\hat{b}_w$).
\begin{equation}
    L_r = \left( b_h - \hat{b}_h\right)^2 + \left( b_w - \hat{b}_w\right)^2.
\end{equation}
During training, we define the ground-truth height and width as the bounding box centered at the object in the first frame that contains the object in the following 7 frames (the other frames in the clip to be processed). This ensures that the object of interest will be present in all frames being processed, even if there is a large amount of motion.

In and end-to-end fashion, we train our network with an objective function which is the sum of these three losses: 
\begin{equation}
    L = L_s + L_D + L_r.
\end{equation}

\section{Experiments}
\paragraph{Datasets}
We evaluate our method on two video object segmentation datasets: Youtube-VOS \cite{xu2018youtube2} and DAVIS-2017 \cite{pont20172017}. Youtube-VOS contains 4,453 videos - 3,471 for training, 474 for validation, and 508 for testing. The training and validation videos have pixel-level ground truth annotations for every 5th frame (6 fps). The DAVIS-2017 dataset contains a total of 150 videos - 60 for training, 30 for validation, 60 for testing. These testing videos are split into a test-dev and test-challenge set, each with 30 videos; we evaluate our method on the test-dev set. The videos in DAVIS-2017 have annotations for all frames. Both datasets contain a wide variety of objects and both contain videos with multiple object instances.

\begin{figure*}[ht!]
\begin{center}
\includegraphics[width=1.0\linewidth]{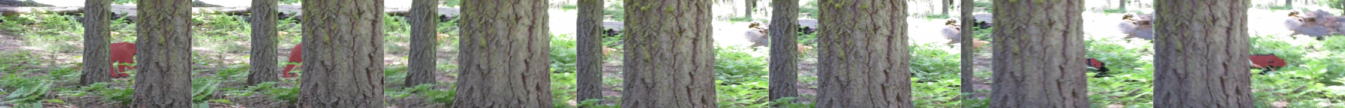}
\includegraphics[width=1.0\linewidth]{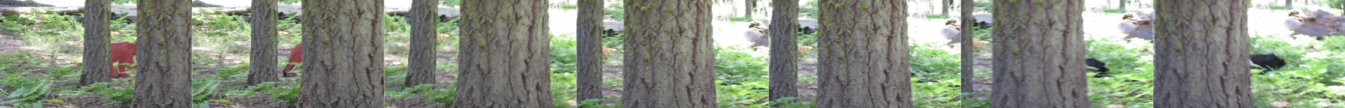}
\includegraphics[width=1.0\linewidth]{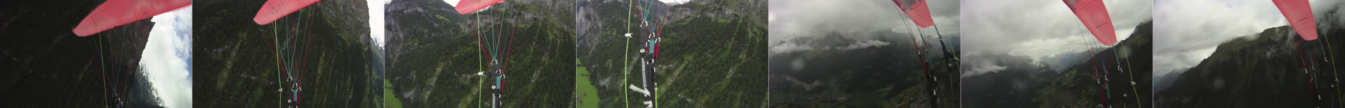}
\includegraphics[width=1.0\linewidth]{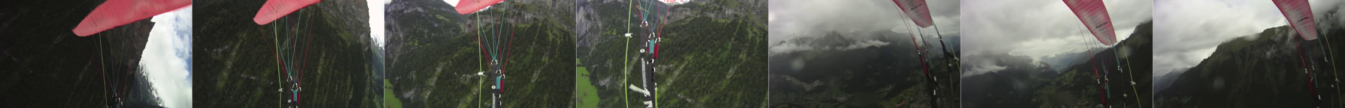}

\end{center}
   \caption{A qualitative comparison between networks with and without the memory module. Rows 1,3: with memory module. Rows 2,4: without memory module. The first example contains a bear which is completely occluded for over 40 frames, but the memory module allows the network to segment the bear when it reappears. The second video shows that the memory module can handle cases where an object leaves and reenters the scene.}
   %The first example shows that the memory module allows the network to overcome the partial occlusion of the van - without the memory module, the segmentation of the van degrades as the human walks in front of it. 
\label{fig:qualitative-occ}
\end{figure*}

\begin{figure*}[ht!]
\begin{center}
\includegraphics[width=1.0\linewidth]{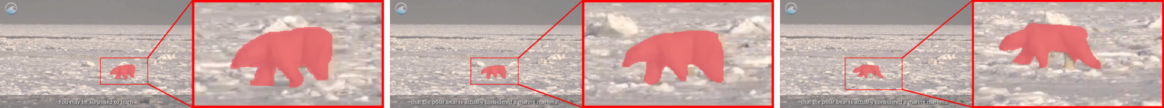}
\includegraphics[width=1.0\linewidth]{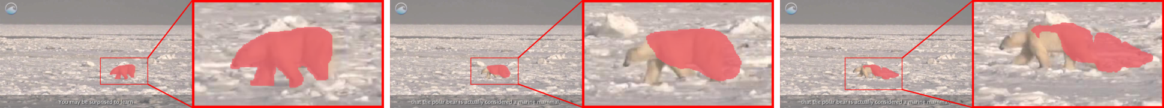}
\includegraphics[width=1.0\linewidth]{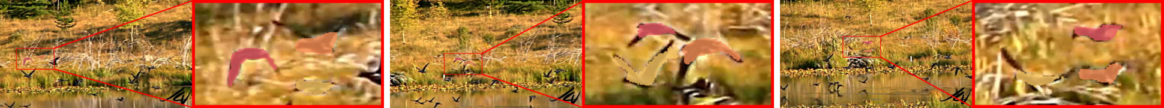}
\includegraphics[width=1.0\linewidth]{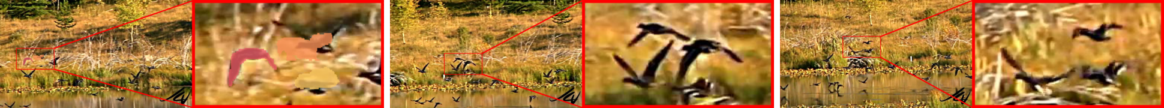}
\includegraphics[width=1.0\linewidth]{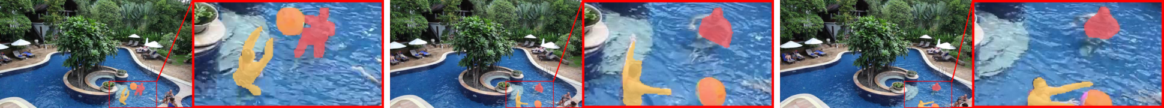}
\includegraphics[width=1.0\linewidth]{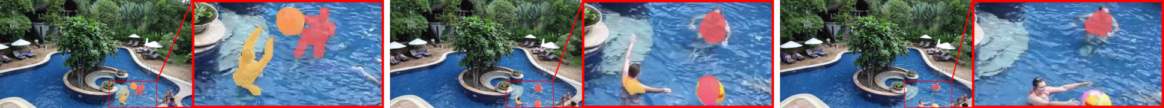}

\end{center}
   \caption{A qualitative comparison between networks with and without the zooming module. Rows 1,3,5: with zooming module. Rows 2,4,6: without zooming module. The first example demonstrates the network's ability to generate fine-grained segmentations on small objects when the zooming module is used.  Very small objects that move rapidly, like those in examples 2 and 3, are lost rather quickly when the zooming module is not present.}
\label{fig:qualitative-bbox}
\end{figure*}

\begin{table*}[ht!]
    \centering
    \begin{tabular}{l|c|c|c|c|c|c|c}
        Method & OL & $\mathcal{J}$ seen & $\mathcal{J}$ unseen & $\mathcal{F}$ seen & $\mathcal{F}$ unseen & Overall & Speed (frames/s) \\
        \hline
        OSVOS \cite{caelles2017one} & \cmark & 59.8 & {\bf 54.2} & 60.5 & {\bf 60.7 } & 58.8 & 0.10 \\
        OSMN \cite{yang2018efficient} & \xmark & 60.0 & 40.6 & 60.1 & 44.0 & 51.2 & 7.14 \\
        S2S (offline) \cite{xu2018youtube} & \xmark & 66.7 & 48.2 & 65.5 & 50.3 & 57.6 & 6.25 \\
        \hline
        {\bf Our Method} & \xmark &  {\bf 67.3} & 53.7 & {\bf 68.1} & 59.9 & {\bf 62.3} & {\bf 13.5}\\
        %Our Method (online) & 65.3 & 53.2 & 69.2 & 61.2 & 62.2 \\
    \end{tabular}
    \caption{Our results on the Youtube-VOS validation set. ``OL" denotes online learning. We compare with OSVOS \cite{caelles2017one} and methods which do not perform online learning.}
    \label{tab:youtubevos}
\end{table*}

\begin{table}[ht!]
    \centering
    \begin{tabular}{l|c|c|c}
         & OSVOS \cite{caelles2017one} & DyeNet \cite{li2018video} & Ours \\
        \hline
        Online Learning & \cmark  & \xmark & \xmark \\
        \hline
        $\mathcal{J}$ Mean $\uparrow$ & 47.2 & 60.2  & 47.4 \\
        $\mathcal{J}$ Recall $\uparrow$ & 50.8 & -  & 54.1 \\
%        $\mathcal{J}$ Decay $\downarrow$ & & & & \\
        \hline
        $\mathcal{F}$ Mean $\uparrow$ & 53.7 & 64.8  & 55.2 \\
        $\mathcal{F}$ Recall $\uparrow$ & 57.8 & - & 64.6 \\
%        $\mathcal{F}$ Decay $\downarrow$ & & & & \\
        \hline
        Global Mean & 50.5 & 62.5 & 51.3 \\
    \end{tabular}
    \caption{Our results on the DAVIS-2017 test-dev set. We compare with OSVOS \cite{caelles2017one} and the offline version of DyeNet \cite{li2018video}}
    \label{tab:davis}
\end{table}

\paragraph{Training} 
The network is trained using the objective function described in \ref{section:objective}. Since our segmentation loss requires segmentations for all 8 frames given to the network and the Youtube-VOS training set contains segmentations every 5th frame, we use the method found in \cite{niklaus2017video} to interpolate the segmentation frames that are unavailable. Training is done using the Adam optimizer \cite{kingma2014adam}, starting with a learning rate of $0.0001$. When training on Youtube-VOS, the method converges in about 400 epochs. For our experiments on DAVIS-2017, we fine-tune the network for an extra 200 epochs on the DAVIS-2017 training videos.

\paragraph{Inference} 
During inference, longer videos are processed one clip (8 frames) at a time; the segmentation generated from one clip is used as the input segmentation for the subsequent clip. We find that having frame overlaps between these clips results in improved segmentations at test time, with only a minor decrease of inference speed. All reported results (both accuracy and speed) use an overlap of 3 frames.

%\paragraph{Online Learning} 
%Following recent works, 
%We also fine-tune our network using the first frame of each test video. Given the first frame and its corresponding segmentation $(x_0, y_0)$, we perform affine transformations to obtain a clip of frames and segmentations $\{(x_0, y_0), (x_1, y_1), ..., (x_7, y_7)\}$ with which we train our network. For each object in the first frame, we train the network on $300$ samples.

\paragraph{Evaluation Metrics}
For both datasets, we evaluate the segmentation results using the region similarity $\mathcal{J}$ and the contour accuracy $\mathcal{F}$ as described in \cite{perazzi2016benchmark}. For Youtube-VOS, results are averaged over the ``{\it seen}" categories - those objects found in training videos - and ``{\it unseen}" categories - the objects present in the validation and testing sets but not present in the training set.

\subsection{Comparison with State-of-the-art}
Since our method does not use online learning, we compare with only offline approaches. The exception to this is OSVOS \cite{caelles2017one}, which is a standard benchmark video object segmentation approach.

\paragraph{Youtube-VOS}
The performance of our network on Youtube-VOS are shown in Table \ref{tab:youtubevos}. Overall, our model performs at least $4\%$ better than all offline methods and $3.5\%$ better than OSVOS. OSVOS slightly outperforms us on unseen categories, but our network has a substantial $8\%$ improvement in both of the ``seen" metrics. Some qualitative results on Youtube-VOS videos are shown in Figure \ref{fig:qualitative-good}.

\paragraph{DAVIS-2017}
Our performance on the DAVIS-2017 test-dev set are shown in Table \ref{tab:davis}. We find that our offline network is unable to achieve better results than many contemporary methods because many of the objects found in DAVIS-2017 do not appear in the Youtube-VOS training set. DyeNet \cite{li2018video} is able to outperform our network by a wide margin; we attribute this to the fact that the method is image based, which allows their region-proposal network and feature extraction network to be pretrained on larger image datasets.

\paragraph{Speed Analysis}

\begin{figure}[ht!]
\begin{center}
\includegraphics[width=1.0\linewidth]{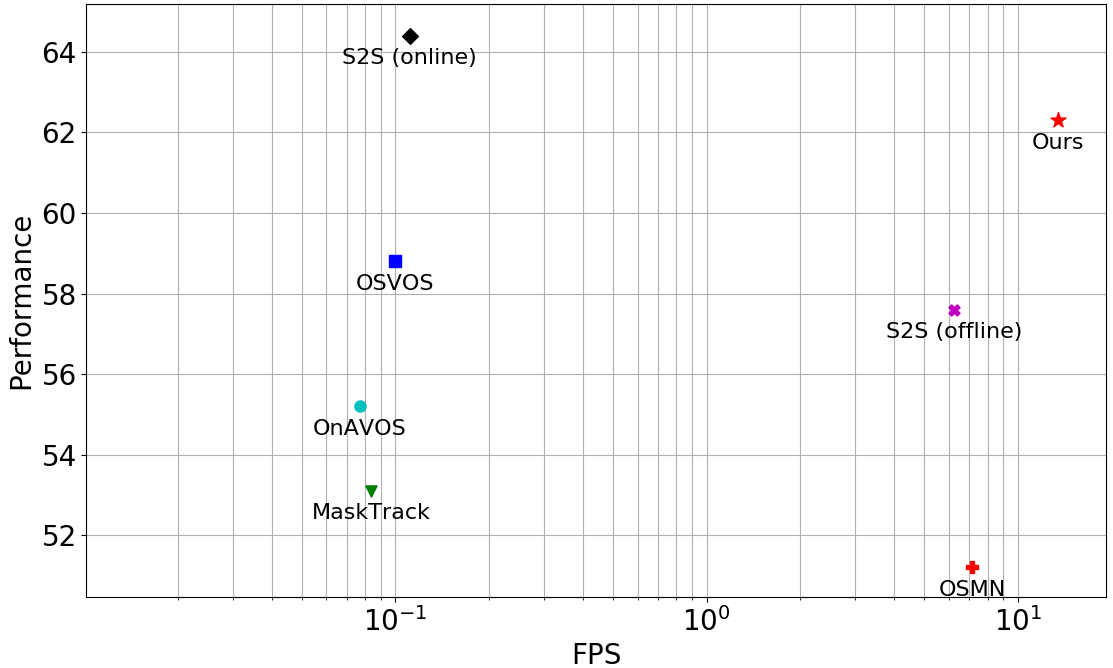}

\end{center}
   \caption{Comparison of quality and speed of previous video object segmentation methods on the Youtube-VOS dataset. We graph the overall performance percentage vs the frames-per-second. The x-axis (fps) is in the log scale.}
\label{fig:speed}
\end{figure}

\begin{table*}[ht!]
    \centering
    \begin{tabular}{l|c|c|c|c|c}
        Ablation & $\mathcal{J}$ seen & $\mathcal{J}$ unseen & $\mathcal{F}$ seen & $\mathcal{F}$ unseen & Overall \\
        \hline
        No Zooming & 62.1 & 45.8 & 61.3 & 48.1 & 54.3 \\
        HC Zooming  & 65.8 & 51.7 & 66.5 & 57.5 & 60.4 \\
        \hline
        Concat Routing & 65.2 & 51.0 & 65.6 & 56.9 & 59.7 \\
        Fully Conv & 64.5 & 51.5 & 64.8 & 57.0 & 59.4 \\
        \hline
        No Memory & 64.9 & 49.6 & 65.3 & 53.9 & 58.4 \\
        \hline
        Full Method &  67.3 & 53.7 &  68.1 & 59.9 & 62.3 \\ 
    \end{tabular}
    \caption{Our ablation experiment results on the Youtube-VOS validation set. Each row corresponds to a different ablation. The final row contains the results of our method without any changes.}
    \label{tab:ablations}
\end{table*}

Running on a Titan X Pascal GPU, our network segments an average of $13.5$ frames per second. We compare our network's inference speed with other approaches in Figure \ref{fig:speed}. Our network is able to segment frames at a much faster rate than previous methods, because we simultaneously segment 8 frames at once as opposed to one frame at a time.

\subsection{Ablation Study}
All ablation experiments are performed on the Youtube-VOS dataset. The quantitative results for the ablations are shown in Table \ref{tab:ablations}.

\paragraph{Zooming Module} To test the effectiveness of our zooming module, we first evaluate our method without any zooming. In this experiment, we resize all frames to $128 \times 224$ and segment them with CapsuleVOS. Without the zooming module, the network's performance decreased by about $8\%$. The zooming module improves the segmentations in two ways: (1) the network is able to keep track of smaller objects, and (2) the network can generate finer segmentation masks for medium sized objects. Figure \ref{fig:qualitative-bbox} shows examples of our method with and without the zooming module; there is a noticeable decrease in segmentation accuracy for smaller objects without the zooming module. We also test if if a simple, hand-crafted zooming method would perform as well as our zooming module. In this experiment, we use a hand-crafted bounding-box around the foreground object in lieu of the zooming module. We find that the hand-crafted bounding-box results in improved segmentations when compared to no zooming, but the zooming module's learned bounding-boxes perform best.

\paragraph{Attention Routing} We run two ablations to test the effectiveness of our proposed capsule routing algorithm. The first is performing conventional EM-routing by simply concatenating the video and frame capsules; the second is removing capsules entirely, and having a fully convolutional network with a similar number of parameters. We find that our proposed routing algorithm does improve segmentations when compared to simple capsule concatenation; this is because the proposed routing algorithm conditions the video capsules based on their agreement with the frame capsules, whereas concatenation does not differentiate between frame and video capsules and attempts to find agreement between all capsules. We also find that the network without capsules performs similar to the network with capsule concatenation; this suggests that the standard EM routing algorithm cannot effectively perform the conditioning operation which this tasks requires and that our proposed routing procedure successfully conditions the video capsules based on the frame capsules.

\paragraph{Memory Module} In this final ablation, we test the importance of the memory module in the frame network. We find that this ConvLSTM improves results by $4\%$, because it allows the network to handle issues like occlusion or when the object of interest leaves the frame. Figure \ref{fig:qualitative-occ} contains some qualitative results depicting the two issues that the memory module solves: occlusion and objects leaving the frame. Once the occlusion ends or the object re-enters the frame, the ConvLSTM allows the network to remember the object which it must segment.

\section{Conclusion}

We have proposed a video capsule network, CapsuleVOS, for semi-supervised video object segmentation. The use of capsules provides an effective modeling of entities present in the video and the attention-based routing helps in the tracking and segmentation of objects. The network contains two additional novel components: a zooming module and a memory module. The zooming module ensures the capture of small objects present in the video and the memory module tracks objects in scenarios when they are occluded or when they move out of the scene. The experimental evaluation demonstrates the effectiveness of our proposed network in video object segmentation and its ability to segment small and occluded objects. Moreover, our ablations show the effectiveness of our proposed routing procedure when compared to the exists EM routing algorithm. The network segments multiple frames at once which allows it to perform segmentation at a much faster rate when compared with existing methods.

%-------------------------------------------------------------------------
\subsection{Acknowledgement}
This research is based upon work supported by the Office of the Director of National Intelligence (ODNI), Intelligence Advanced Research Projects Activity (IARPA), via IARPA R\&D Contract No. D17PC00345. The views and conclusions contained herein are those of the authors and should not be interpreted as necessarily representing the official policies or endorsements, either expressed or implied, of the ODNI, IARPA, or the U.S. Government. The U.S. Government is authorized to reproduce and distribute reprints for Governmental purposes notwithstanding any copyright annotation thereon.

{\small
\bibliographystyle{ieee_fullname}
\bibliography{egbib}
}

\end{document}